\documentclass[conference,letterpaper,10pt]{ieeeconf}
\usepackage[left=1.91cm,right=1.91cm,top=1.91cm,bottom=1.91cm]{geometry}

\usepackage{graphicx}
\usepackage{amssymb}
\usepackage{amsmath}
\usepackage{cite}
\usepackage{comment}
\usepackage{color}
\usepackage{url}
\usepackage{alltt}
\usepackage{cleveref}
\usepackage{tikz}
\usepackage{pgfplots}
\usepackage{pgf-umlsd}
\usepackage{cprotect}
\usepackage{multirow}
\usepackage{algorithm}
\usepackage[noend]{algpseudocode}
\usepackage{bigfoot}
\usepackage{paralist}
\usepackage{tabularx}
\usepackage[whole]{bxcjkjatype}
\usepackage{ifthen}
\usepackage{threeparttable}
\usepackage[hang,small,bf]{caption}
\usepackage[subrefformat=parens]{subcaption}
\usepackage[super]{nth}
\usepackage{colortbl}
\usepackage{scalefnt}
\usepackage{threeparttable}
\usepackage{pifont}

\pgfplotsset{compat=1.14}
\IEEEoverridecommandlockouts

\makeatletter
\newcommand{\figcaption}[1]{\def\@captype{figure}\caption{#1}}
\newcommand{\tblcaption}[1]{\def\@captype{table}\caption{#1}}
\makeatother

\title{\LARGE \bf
Sim-to-Real Transfer of Compliant Bipedal Locomotion on\\Torque Sensor-Less Gear-Driven Humanoid
}

\author{
Shimpei Masuda$^{1, 2}$,
Kuniyuki Takahashi$^{1}$
\thanks{
        $^{1}$S. Masuda and K. Takahashi are with Preferred Networks, Inc.
        {\tt\footnotesize
        \{{masuda, takahashi\}@preferred.jp}}
        $^{2}$S. Masuda is also with the University of Tsukuba. This work is done in Preferred Networks, Inc.
}
}

\begin{document}

\setlength\floatsep{3.0mm}
\setlength\textfloatsep{3.0mm}
\setlength\intextsep{0pt}
\setlength\abovecaptionskip{0pt}
\setlength{\belowdisplayskip}{0pt}

\maketitle
\thispagestyle{empty}

\begin{abstract}
Sim-to-real is a mainstream method to cope with the large number of trials needed by typical deep reinforcement learning methods.
However, transferring a policy trained in simulation to actual hardware remains an open challenge due to the \emph{reality gap}.
In particular, the characteristics of actuators in legged robots have a considerable influence on sim-to-real transfer.
There are two challenges:
1) High reduction ratio gears are widely used in actuators, and the \emph{reality gap} issue becomes especially pronounced when backdrivability is considered in controlling joints compliantly.
2) The difficulty in achieving stable bipedal locomotion causes typical system identification methods to fail to sufficiently transfer the policy.
For these two challenges, we propose 1) a new simulation model of gears and 2) a method for system identification that can utilize failed attempts.
The method's effectiveness is verified using a biped robot the ROBOTIS-OP3, and the sim-to-real transferred policy can stabilize the robot under severe disturbances and walk on uneven surfaces without using force and torque sensors.
\footnote{An accompanying video is available at the following link:\\ \url{https://www.youtube.com/watch?v=-QHx5V9oZDc}}
\end{abstract}
\section{Introduction}
\label{sec:introduction}
If robots are able to achieve stable bipedal walking, they can traverse numerous challenging terrains, such as uneven roads and stairs.
To achieve this mission, robots need to be robust against disturbances and changes in environments.

Deep Reinforcement Learning (DRL) has been gaining attention as an appealing alternative for high-performance robot control for situations where manually designed alternatives are difficult in the field of locomotion control~\cite{Lee2020quadwild, Siekmann2021cassie-blind, Peng2020imitation}.
Since DRL methods typically require an inordinate number of trials, the sim-to-real approach, in which policies are first trained using simulations and then transferred to a real-world robot, has been attracting attention~\cite{tan2018simtoreal, du2021autotuned, Yu2019PUP}.
However, the limited fidelity of simulations leads to differences in their behavior compared to the real world.
This gap is referred to as the reality gap and hinders policies trained in simulation from directly being applicable in the real environment.

\begin{figure}[t]
  \centering
  \includegraphics[width=0.90\columnwidth]{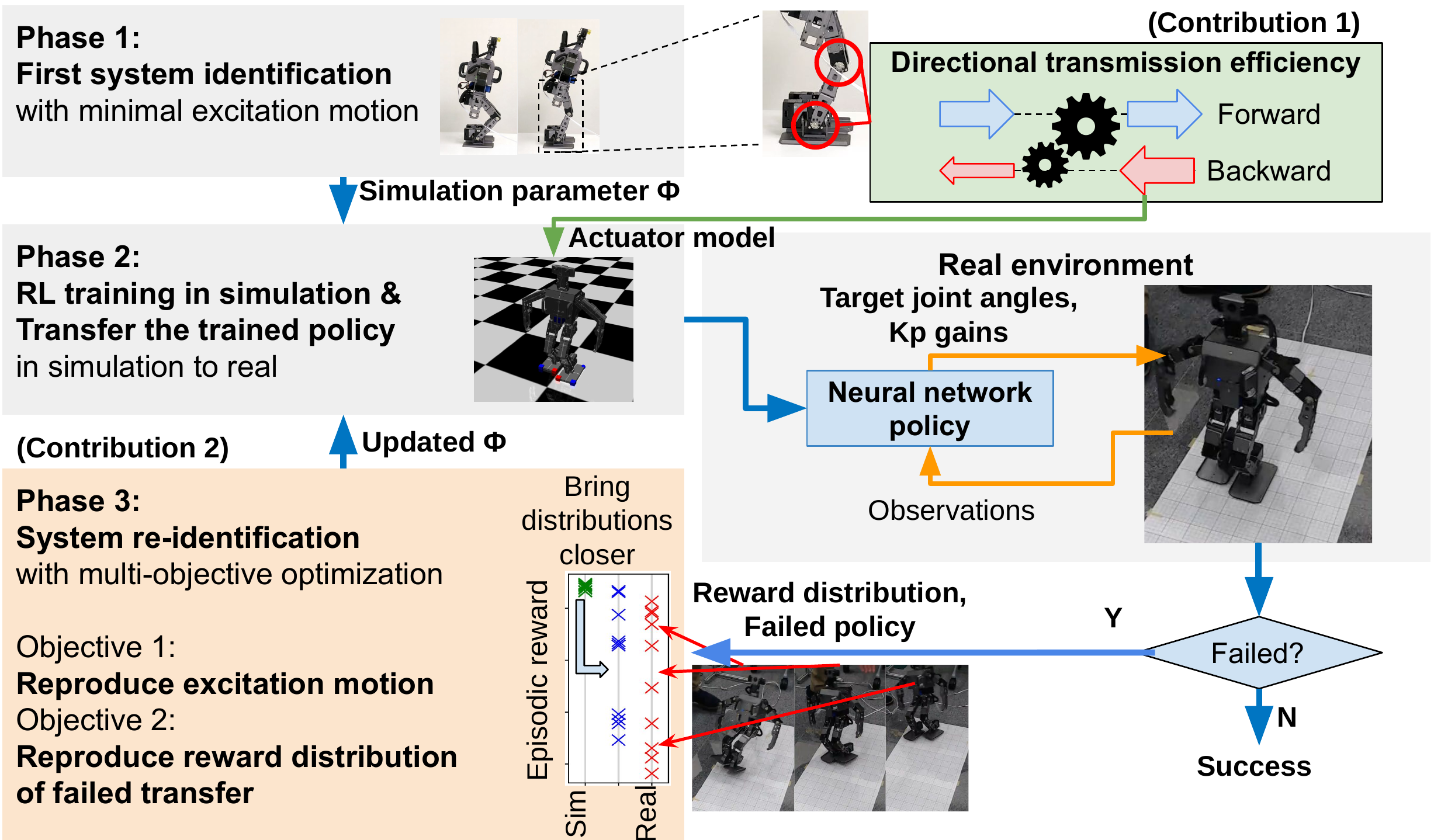}
  \caption{Concept of the proposed methods}
  \label{fig:detailed_overview}
  \vspace{-4mm}
\end{figure}

There are two popular approaches to narrow this reality gap; making simulations more realistic and making control policy robust against model errors using methods such as Domain randomization~\cite{tobin2017domain}.
We focus on the former because the fine-grained behavior of the actuator needs to be mastered by the policy to achieve superior performance, and domain randomization may not be good at this.
There are methods to make a realistic simulation by identifying simulation parameters such as friction and inertial parameters of the physics model reproducing the actual robot behavior~\cite{chebotar2019closing, du2021autotuned, tan2018simtoreal, Xie2020cassie-simtoreal, Hwangbo2019actuatornet, Yu2019PUP}.
Especially for legged robots, it has been reported that the model of the actuators has a more significant impact on the reality gap~\cite{Hwangbo2019actuatornet, Yu2019PUP, Xie2020cassie-simtoreal}.
However, the analytical actuator models used in previous studies do not adequately reproduce the characteristics of transmission efficiency, particularly in gears with high reduction ratios~\cite{albert2015directional, Matsuki2019backdrive}.
This characteristic of transmission efficiency strongly influences backdrivability, the passive movement of joints, due to external/reactive forces from the environment.
Moreover, this backdrivability is important for improving locomotion performance~\cite{Hyon2006passivity, Mesesan2019passivity, Suzuki2017sensorless-torque}.
On the other hand, a configuration that uses a high reduction ratio gear without a joint torque sensor is still a popular option in robots that require low-cost and high power, such as humanoids.
Therefore, it is useful if DRL can control sensor-less gear-driven actuators by exploiting backdrivability, which was difficult to design manually due to the non-linear characteristics of gears.
In this paper, we propose a physics simulation implementation model of gear-driven actuators and train the control policy using DRL.
The gear simulation model has only a few parameters and is compatible with the system identification described below.

To make simulations more realistic, approaches to building a model of the physics engine representing the robotic system and then identifying the various parameter values of the model have been widely attempted and are often referred to as system identification.
To properly identify the system parameters in the domain of the target task, we need to carefully design the data characteristics, such as ranges of joint velocity or torque to be collected or collect large amounts of data, but it increases the risk of robot failure.
To improve the efficiency of data collection, using the data while task execution is considered promising.
A possible pipeline would be first to perform a rough system identification, then train policy and transfer once, collect the data on which the policy performs the task, and identify simulation parameters again.
In this paper, we call the second system identification as system re-identification.
However, re-identification by matching time-series sensor data, as done in existing research~\cite{chebotar2019closing}, is difficult for tasks with a narrow set of stable states, such as bipedal balancing and walking, because a slight difference in action can easily make a big difference in the future state compared to tasks where the robot arms are fixed to a base.
This paper proposes a re-identification method that focuses on acquired rewards as an abstract value for actual and simulated behavior and uses the difference as a proximity index for behavior.

To summarise the challenges, there is the reality gap problem in applying sim-to-real transfer to torque sensor-less and gear-driven robots with bipedal locomotion tasks, and the following two factors contribute to this challenge.
\begin{enumerate}
    \item Commonly used actuator models in simulations are inadequate in modeling high reduction ratio gears.
    \item A slight identification error can lead to a massive failure of the sim-to-real transfer, and re-identification by matching time-series sensor data is difficult.
\end{enumerate}
In this study, we propose the following to address these issues to reduce the reality gap and increase the success rate of sim-to-real transfers,
\begin{enumerate}
    \item An actuator model of the high reduction ratio gears for physics simulation.
    \item A new simulation parameters identification method that utilizes the data on failed transfer, using the acquired rewards in the actual robots.
\end{enumerate}
And we show the effectiveness of the proposed method by realizing robust bipedal locomotion, including balancing and walking on an actual robot.

\section{Related work}
\label{sec:related works}
\subsection{Joint control in RL for legged robots}
In recent years, many studies on Reinforcement Learning (RL) for legged robots have been conducted, and many of them have achieved high performance on real robots~\cite{tan2018simtoreal, Xie2020cassie-simtoreal, Siekmann2021commongaits}.
These studies often applied a position controller to each joint, and an RL policy commands target angles.
One of the reasons is that position control improves learning efficiency and task performance compared to torque or velocity control~\cite{Peng2017actionspace}.
On the other hand, some studies have been conducted to improve robustness to disturbances and environmental recognition errors by exploiting passivity (in other words, utilizing the joint backdrivability)~\cite{Hyon2006passivity, Mesesan2019passivity, Suzuki2017sensorless-torque}.
Even position control can have backdrivability by specifying a low gain, but utilizing that is not well-explored in sim-to-real studies.
In this study, we focus on exploiting backdrivability in robots with gear-driven actuators and address some of the difficulties described in the following subsections.

\subsection{Sim-to-real RL for Robotics}
It is widely known that controllers trained by RL in simulations do not work well in actual robots due to the reality gap, and various approaches have been developed to overcome this challenge.
Domain randomization~\cite{tobin2017domain} is a frequently used approach to compensate for the reality gap.
The method trains an RL policy in many simulations with randomized dynamics, sensors, or appearances, such as joint friction or textures of the object.
Domain randomization works by increasing the robustness of the controller to a wider range of configurations in the hopes of better transferring to the real world.
However, specifying appropriate distributions of randomization for the various simulation parameters is a difficult problem.
Too much randomization tends to make the RL policy too conservative, turning the performance poor, and too little will not have the desired generalization effect.

Another approach to address the reality gap is precisely reproducing the real behavior of the system on the simulation~\cite{chebotar2019closing, du2021autotuned} and training a policy on that.
Assuming that simulation is characterized by some dynamics model parameters, such as the size of joint friction, reproducing the real behavior corresponds to finding parameter values or distributions.
This process is generally referred to as system identification.
We can roughly categorize prior work that takes this approach as  1) manual system identification~\cite{tan2018simtoreal, Xie2020cassie-simtoreal}, 2) system identification based on collected generic data~\cite{Hwangbo2019actuatornet, Yu2019PUP}, and 3) system identification based on rolled out RL policy behavior ~\cite{du2021autotuned, chebotar2019closing}.
Since our focus is on frictional gear-driven joint and utilization of backdrivability, our approach combines 2) and 3).
If we choose 1) or 2) approach, it may require a large amount of data on situations, including various external torques and driving states.
Moreover, tasks with instability, such as bipedal walking, where a slight difference in action can easily make a big difference in the future compared to tasks with arms fixed to a base.
In such cases, precise system identification is challenging by 3) with trajectory comparison~\cite{chebotar2019closing} or parameter prediction~\cite{du2021autotuned}.
In this paper, we use both 2) and 3) as complementary approaches.
It allows relatively small data collection and system identification using the failed trials of the rolled-out RL policy.

\subsection{Modeling Actuators in Sim-to-Real Transfer}
The reality gap is especially critical for actuators in legged robots.
Several kinds of research have been conducted to improve the actuator models in the simulation, broadly classified into two categories to address the challenge:
1) Use a neural network for the actuator model and train with largely collected data~\cite{Hwangbo2019actuatornet, Yu2019PUP} (Network training corresponds to system identification).
2) Design a detailed model of the actuator and implement it in the simulation~\cite{tan2018simtoreal, taylor2021sonybipedal}

The first method can potentially build highly accurate models but requires large amounts of data.
In this study, we extend category 2) to model actuators with high reduction ratio gears with low-dimensional parameters.
Parameters would be obtained with relatively small amounts of data since the number of parameters of the gear model is small.
In addition, since high reduction ratio gears are a major characteristic, the model can be applied to many actuators.

\section{PRELIMINARY of RL}
\label{sec:preliminary}
Let $(S, A, T, R, \gamma, s_0)$ be a Markov decision process with state $s\in S$, action $a\in A$, transition probability $T(s_{t+1}\vert s_t, a_t)$, reward function $R(s, a)\in \mathbb{R}$, discount factor $\gamma \in [0, 1]$, and initial state distribution $s_0$.
RL typically learns a policy $\pi: S\rightarrow A$ to maximize the expected discounted accumulated reward $\mathbb{E}[\sum_{t=1}^{T}\gamma^{(t-1)}R(s_t,a_t)]$, where $T$ is the episode length.
In our robot control problem, $S$ corresponds to sensor data, $A$ to commands to joint actuators, and state transitions to steps in the control loop.
Although various RL algorithms have been proposed, in this study, we use Soft-Actor-Critic~\cite{haarnoja2018sac}, which supports continuous action spaces and has been reported to have high performance on a wide range of tasks.
Specific action space, observation space, and reward function are described in \Cref{sec:action_obs_reward}.

\section{Method}
\label{sec:proposed_method}
The proposed sim-to-real transfer method consists of two main components: a gear-driven actuator model for physics engines to improve robot simulation, and a system identification method to specify the simulation parameters for RL training.
The method consists of the following three phases.
Phase~1: Perform first system identification with excitation motion, such as squatting motion.
Phase~2: Using the identified parameters, train the RL policy using simulation, then run the trained policy on the actual robot.
Phase~3: If the robot does not work well, re-identify the simulation parameters using the failed attempts.
Phase~2 is then executed again.

We first describe the proposed actuator model.
Then, the first system identification is described, followed by the system re-identification.
\subsection{Actuator Modeling}
\label{sec:Actuator modeling}
\begin{figure}[t]
  \centering
  \includegraphics[width=0.9\columnwidth]{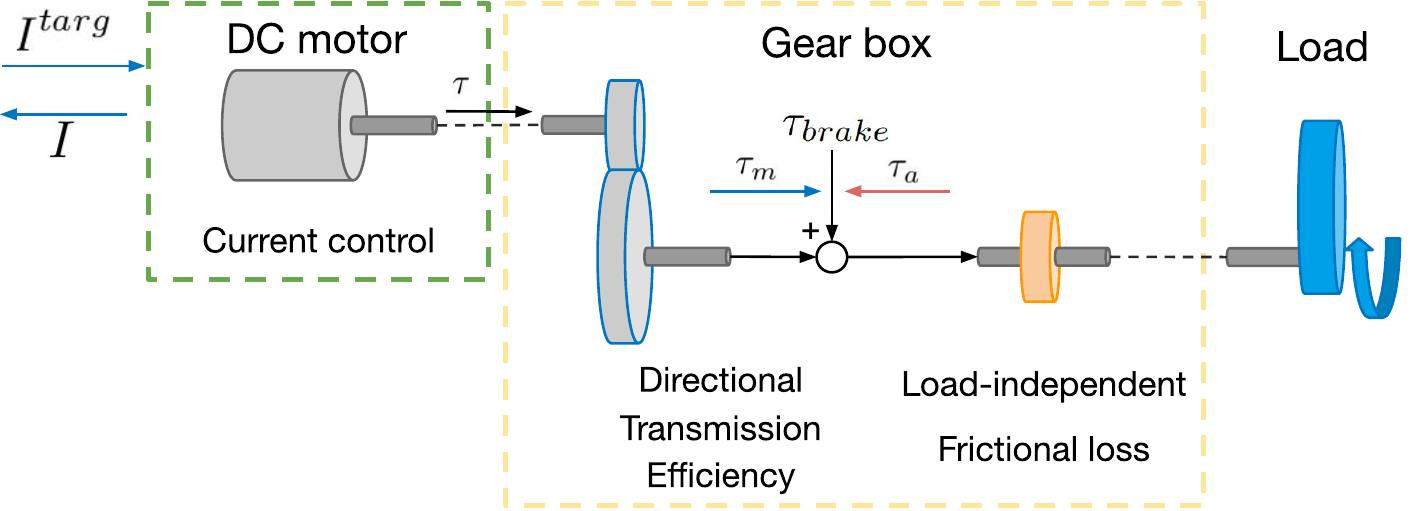}
  \caption{Overview of the actuator modeling}
  \label{fig:actuator_modeling}
  \vspace{-4mm}
\end{figure}

This section describes our actuator model.
The actuator consists mainly of a DC motor (\Cref{sec:DC motor model}) and gears (\Cref{sec:Model of gear friction}), as illustrated in Fig.~\ref{fig:actuator_modeling}.
The actuator is controlled by variable gain PD, as described in \Cref{sec:variable gain PD}.
In this study, we implement this actuator model on the physics engine Mujoco~\cite{Todorov2012mujoco}, but note that it can be implemented in other physics engines.

\subsubsection{DC Motor Model}
\label{sec:DC motor model}
For the DC motor model, we use the commonly used model~\cite{tan2018simtoreal}, which is represented by the following equation:
\begin{equation}
    \label{eq:DCMotor}
    \tau = K_{t} I, \ where \ 
    I = \frac{V_{pwm} - V_{back-emf}}{R_{ter}}, \ 
    V_{back-emf} = K_{t} \dot{q}
\end{equation}
where $\tau$ is the exerted torque by the motor, $I$ is the current, $K_{t}$ is the torque constant of the motor, $R_{ter}$ is the terminal resistance, $V_{pwm}$ is the voltage applied to the motor, $V_{back-emf}$ is the back electromotive force (EMF) voltage, and $\dot{q}$ is the angular velocity of the shaft.

A current controller controls voltage to drive the DC motor with the target current.
The applied voltage $V_{pwm}$ at each simulation time step can be formulated as follows using the target current $I^{targ}$ and battery voltage $V_{battery}$:
\begin{equation}
    \label{eq:CurrentControl}
    \begin{split}
    &V_{pwm}^{targ} = R_{ter} I^{targ} + V_{back-emf}\\
    &V_{pwm} = min(max(V_{pwm}^{targ}, -V_{battery}), V_{battery})
    \end{split}
\end{equation}
\subsubsection{Gear Model with Directional Transmission Efficiency}
\label{sec:Model of gear friction}
There is an asymmetry in transmission efficiency depending on the drive direction~\cite{albert2015directional,Matsuki2019backdrive}, and often this asymmetry is more significant, especially at high reduction ratios.
We propose a simulation model considering this directional transmission efficiency (DTE) characteristic.
The model has the feature of reproducing the loss by transmission efficiency through additional braking torque and is easy to incorporate into the existing physics engine. 

The gear model considers the transmission efficiency and load-independent friction loss.
The former depends on the force, while the latter mainly depends on the angular velocity.
The method treats the force loss due to transmission efficiency as an additional brake torque applied to the joint in each simulation time step.
The brake torque is calculated from the generated torque by the motor, load torque, and the forward and backward transmission efficiencies.
Assuming that the simulation time step is small enough for the target dynamics, the brake torque is calculated as follows:

\footnotesize
\vspace{-2mm}
\begin{eqnarray}
    \label{eq:GearEfficiency}
    \begin{split}
    \tau_{brake}(\tau_{m}, \tau_{a}, \eta_{fw}, \eta_{bw}) =\qquad\qquad\qquad\qquad\qquad\qquad\\
    \left\{
    \begin{array}{ll}
    &-L_{fw} \quad\quad\quad if \quad sign(\eta_{fw} \tau_{m} + \tau_{a}) = sign(\tau_{m})\\
    &-L_{bw} \quad\quad\quad if \quad sign(\tau_{m} + \eta_{bw} \tau_{m}) = sign(\tau_{a})\\
    &-(\tau_{m} + \tau_{a}) \quad else
    \end{array}
    \right.\\
    where, L_{fw} = (1 - \eta_{fw}) \tau_m,\quad L_{bw} = (1 - \eta_{bw}) \tau_{a}
    \end{split}
\end{eqnarray}
\normalsize
where $\tau_{m}$ is the generated torque from the motor, $\tau_{a}$ is the load torque on the joint, and $\eta_{fw}$ and $\eta_{bw}$ are the forward and backward transmission efficiencies, respectively.
$\tau_{m} = r_{gear}\tau$ where $r_{gear}$ is gear reduction ratio.
When $\tau_{m}$ is clearly greater than $\tau_{a}$ or the directions of $\tau_{m}$ and $\tau_{a}$ are the same, it is regarded as a forward drive state, and the brake torque based on the $\eta_{fw}$ is generated.
If the directions of $\tau_{m}$ and $\tau_{a}$ are opposite and $\tau_{a}$ is larger even after considering the $\eta_{bw}$, it is considered to be in the backward drive state, and the brake torque based on the $\eta_{bw}$ is generated.
A state that does not meet either condition is an antagonistic state with no apparent difference between $\tau_{m}$ and $\tau_{a}$, and thus generates a brake torque, such that $\tau_{m} + \tau_{a} + \tau_{brake} = 0$.
Since $\tau_{a}$ is approximated using the value of the current simulation state, there is a concern that the simulation may become unstable.
However, it is expected to be stable when used in conjunction with the load-independent friction as below.
The load-independent friction loss generates torque that cancels joint torque in the absolute upper bound.
In particular, Mujoco will handle it as a constraint and allow the constant value.
So we calculate the friction loss value at each time step to generate approximated static and viscous friction.
For the calculation, Stribeck function ${\tau}_{fric}$ is used:
\begin{equation}
    \label{eq:StribeckFriction}
    {\tau}_{fric} = f_{c} + s(f_{s} - f_{c}) + k_{v}|\dot{q}|, \ where \ 
    s = \frac{exp(|\dot{q}|)}{{\dot{q}}_{static}}\\
\end{equation}
where each parameter $f_{s}$, $f_{c}$, $k_{v}$, and $\dot{q}_{static}$ are the static friction force, the Coulomb friction coefficient, the viscous friction coefficient, and the angular velocity value regarded as non-static, respectively.

\subsubsection{Variable Gain PD}
\label{sec:variable gain PD}
The actuator model described so far is controlled by the PD controller.
We adopted this because previous research has reported that placing the position control at the lower level of the RL policy improves learning efficiency and performance~\cite{Peng2017actionspace}.
Generally, the PD controller is used with fixed gains, but in this study, {\bf the gains are also commanded with the target position}.
It allows a strategy to increase backdrivability while reducing target position tracking performance by specifying a smaller gain, depending on the task situation.

\subsection{System Identification for Sim-to-real Transfer}
\label{sec:model identification for Sim-to-real}
This section describes the system identification method of the simulation parameters, such as the motor torque constant or frictions of the actuator model shown in \Cref{sec:Actuator modeling}.
\subsubsection{First System Identification}
\label{sec:model identification}
First, a suitable motion for system identification (we call it excitation motion) is designed as time series data $\mathbf{\theta^{targ}}$ of the target angle of each joint.
The excitation motion used in this study will be described in \Cref{sec:Sim-to-Real transfer}.
A simple position control feedback system is provided to track $\mathbf{\theta^{targ}}$ in the actual robot.
We then collect a set of sensor data $\mathbf{O}^{real}$ for the robot in motion.
This sensor data includes the angle of each joint, angular velocity, current, and the tilt of the upper body.
Similarly, on the simulation built based on the parameters $\phi$ that includes parameters of body inertial and actuator model (actually used in this study are shown in \Cref{tab:model_paramters}), the same position control feedback system is used to perform the motion to track $\mathbf{\theta^{targ}}$ and obtain $\mathbf{O}^{sim}$.
The parameter $\phi$ that minimizes the difference between $\mathbf{O}^{real}$ and $\mathbf{O}^{sim}$ is calculated by sampling-based black-box optimization as follows:
\vspace{-2mm}
\begin{eqnarray}
    \begin{split}
    & \;\;\; \underset{\phi}{\text{min}}
    &L_{exc}(\mathbf{O}^{sim}(\phi), \mathbf{O}^{real})
    \label{eq:sim_calib_overview}
    \end{split}
\end{eqnarray}
where $L_{exc}$ is an error function calculates distance between $\mathbf{O}^{sim}$ and $\mathbf{O}^{real}$.
The specific definition of $L_{exc}$ is described in \Cref{sec:Sim-to-Real transfer}.
Some related works employ a similar method for system identification~\cite{tan2016simcalib, Yu2019PUP}.
The method is suitable for sim-to-real transfer due to the following advantages: 
a) The simulation can be improved from the results of the actual robot alone without detaching the actuator from the robot, and b) since it is optimizing the various parameters simultaneously, the behavior of the combination of each parameter can be matched to the actual robot which reduces the reality gap.
Note, however, that even when looking for combinations of parameters, characteristic components such as gear models need to be modeled.

\subsubsection{System Re-Identification from Failed Transfer}
\label{sec:model identification from failure motion}
Even if we use the simulation improved by the identification described in the previous section for RL training, the transferred policy sometimes does not work.
For example, the actual robot may not even be able to stand on flat ground if a policy has been trained to balance on the tilting board in a simulation.
In such cases, task-relevant data is lacking in the first system identification, and a possible solution is using data collected by the transferred policy while performing the task.
The goal here is to find simulation parameters that reproduce the behavior of an actual robot not only for the excitation motion but also for the behavior while doing a task by the transferred policy.
However, especially in bipedal tasks, a slight difference in policy command can easily make a big difference in the future state, and the sensor data, such as joint angle trajectory, widely diverge.
So, it is challenging to properly calculate the distance between behaviors on simulation and actual by the transferred policy (we call it a policy performance gap) by comparison of sensor data for each time series in the same way as eq.~\eqref{eq:sim_calib_overview}.
Therefore, we employ the acquired reward as an abstract value for the behavior and calculate the distance between two empirical distributions of the reward of episodes as a proximity index.
We formulate this re-identification process as a multi-objective optimization problem with two objectives: the minimization of the excitation motion gap to achieve accurate identification and the minimization of the policy performance gap.
\begin{eqnarray}
    \begin{split}
    & \;\;\; \underset{\phi}{\text{min}}
    &L_{exc}(\mathbf{O}^{sim}(\phi), \mathbf{O}^{real}), W(\mathbf{R}^{sim}(\phi), \mathbf{R}^{real}) \label{recalib_overview}
    \end{split}
\end{eqnarray}
Where, $\mathbf{R}^{sim}$ and $\mathbf{R}^{real}$ are a list of cumulative rewards of each $K$ time task episode.
$\mathbf{R} = [R_{k=1}, R_{k=2} ... R_{k=K}]$ and $R_{k} = \sum_{t=1}^{T}R_{eval}(s_t,a_t)$.
$R_{eval}$ is a reward function for evaluation that is used for both real results and simulation.
$\mathbf{R}^{sim}(\phi)$ is the reward values for running policy $\pi$ in $K$ times on the simulation parameterized $\phi$.
$\mathbf{R}^{real}$ is the reward of episodes in the actual environment, and it will be fixed during optimization.
$W$ calculates the Wasserstein distance for two empirical distributions of rewards.

Optimization provides simulation parameters $\phi$ that make the acquired reward distribution as close to reality as possible, keeping the error in the excitation motion as small as possible. 
In the ideal problem setting, the first objective $L_{exc}$ and the second objective $W$ can be optimal simultaneously, and multi-objective optimization is unnecessary. However, in the real problem, $L_{exc}$ and $W$ are in a bit trade-off relationship, so multi-objective optimization suits this method.

\section{Experimental setup}
\label{sec:experimental setup}
\subsection{Robot Setup: ROBOTIS-OP3}
\label{sec:robot setup}
In this study, we use ROBOTIS-OP3~\cite{ROBOTIS-OP3}.
ROBOTIS-OP3 has a length of $51\mathrm{\,cm}$, a weight of about $3.5\mathrm{\,Kg}$, 6-DOFs in each leg, and 20-DOFs in total.
All the joints of the robot are composed of Dynamixel servo motors XM430-W350.
The stall torque is $4.1\mathrm{\,Nm}$ at $12.0\mathrm{\,V}$ and $2.3\mathrm{\,A}$.
Metal spur gears are used for gears and have a reduction ratio of 353.5.
For the PD control of the joints, the control loop was executed by a PC on the robot, not in servos, and the output value of PD is commanded to the servo as the target current.
This configuration is intended to avoid unknown specifications inside the servo, such as gain calculation, as much as possible.
Due to communication constraints, the frequency of PD control is $125\mathrm{\,Hz}$.
Action inference of the policy is performed on a PC external to the robot at $31.25\mathrm{\,Hz}$.
Note that the communication latency is considered in our simulation.

\begin{table}[t]
    \centering
    \caption{List of parameters for system identification}
    \begingroup
    \scalefont{0.90}
    \begin{tabular}{c|c|c}
        \hline
        Parameter & Unit & Range \\
        \hline \hline
        motor Kt & - & [0.003, 0.009] \\
        motor $R_{ter}$ & $\Omega$ & [4.0, 9.0] \\
        motor armature & $kgm^2$ & [0.0025, 0.011] \\
        gear forward efficiency & - & 1.0 (Fixed) \\
        gear backward efficiency & - & [0.6, 1.0] \\
        joint $f_c$ & Nm & [0.01, 0.25] \\
        joint $k_v$ & - & [0.0025, 0.15] \\
        joint $f_s$ & Nm & $f_c$ + [0.0, 0.25] \\
        base mass offset & Kg & [0.0, 0.5] \\
        base CoM offset x & m & [-0.02, 0.02] \\
        base CoM offset z & m & [-0.02, 0.02] \\
    \end{tabular}
    \endgroup
    \label{tab:model_paramters}
    \vspace{-2mm}
\end{table}
\subsection{System Identification}
\label{sec:Sim-to-Real transfer}
Simulation parameters used in this study for system identification are shown in Table~\ref{tab:model_paramters}.
Note that the motor armature is the additional link inertia caused by the rotor inertia of the DC motor amplified with the gear reduction ratio.
The simulation was implemented using Mujoco~\cite{Todorov2012mujoco}.
The size of the simulation time step was set to $1\mathrm{\,ms}$.

For the eq.~\eqref{eq:sim_calib_overview} of the black-box optimization for first system identification, we use the Tree-structured Parzen Estimator algorithm implemented in Optuna~\cite{Akiba2019optuna} since it gives better results than CMAES in our experimental setup.
For the multi-objective optimization eq.~\eqref{recalib_overview} for re-identification, we used the NSGA-II algorithm, which is known to perform well in multi-objective optimization provided implementation by Optuna.
We set the number of simulation trials for each optimization in the experiment to 2,000 for the first system identification and 3,000 for re-identification.

A simple squatting motion is used for the excitation motion $\mathbf{\theta}^{targ}$ for system identification.
While keeping a forward-leaning posture, execute a flexion/extension movement such that the knee joint angle flexes from $1.47\mathrm{\,rad}$ to $0.6\mathrm{\,rad}$ at a speed of approximately $0.5\mathrm{\,Hz}$.
For the function $L_{exc}$ to evaluate the difference of motions, eq.~\eqref{eq:IdenLossFunc} is used.

\vspace{-2mm}
\scriptsize
\begin{align}
    \label{eq:IdenLossFunc}
    &L_{exc} = \frac{1}{T}\sum_{t=1}^T (
        ||r^{sim}_{t} - r^{real}_{t}||^2 + ||\dot{r}^{sim}_{t} - \dot{r}^{real}_{t}||^2 ) +\\
    &\frac{1}{NT}\sum_{t=1}^T\sum_{i=1}^N (
        ||\theta^{sim}_{t,i} - \theta^{real}_{t,i}||^2+
        ||\dot{\theta}^{sim}_{t,i} - \dot{\theta}^{real}_{t,i}||^2 +
        ||I^{sim}_{t,i} - I^{real}_{t,i}||^2 ) \nonumber
\end{align}
\normalsize
Where, $r$ is upper body orientation, $\dot{r}$ is body angular velocity, $\theta$ is the joint angle, and $I$ is joint current.
$N$ is the number of focusing joints, and $T$ is the number of time steps of the motion.

\subsection{Task for Evaluation}
\label{sec:task}
We designed two tasks for method evaluation, the balancing task and the walking task.
For both tasks, we use common action space, observation space, and reward function.
The common parts are described first, followed by a detailed description of each task.

\subsubsection{Action space, observation space, and reward function}
\label{sec:action_obs_reward}
The action of the policy is the target angle and P gain of PD control for each joint.
There are $10$ joints to be controlled, five in each leg, and the action space has 20 dimensions.
Since our experimental setup doesn't need explicit yaw axis rotation for task completion, we omit the hip-yaw axis for simplification. 
The P gain is commanded at a value between 6 and 0.1, and the D gain is fixed at 0.1.
When the gain is at a small value, the joint is compliantly backdriven by external forces in the vicinity of the target angle.

\begin{table}[t]
    \centering
    \caption{Coefficients of reward function}
    \begingroup
    \scalefont{0.85}
    \begin{tabular}{c|c|c|c}
        \hline
        \begin{tabular}{c}
          Coefficients
        \end{tabular} &
        \begin{tabular}{c}
          Balancing
        \end{tabular} &
        \begin{tabular}{c}
          Walking
        \end{tabular} &
        \begin{tabular}{c}
          $R_{eval}$
        \end{tabular} \\
        \hline\hline
        $K_{bipedal}$ & 0 & 0.4 & 0 \\
        $K_{cmd}$ & 0.6 & 0.3 & 0.6\\
        $K_{smooth}$ & 0.1 & 0.1 & 0.1 \\
        $K_{xd}$ & 2 & 15 & 2 \\
        $K_{yd}$ & 2 & 1 & 2 \\
    \end{tabular}
    \endgroup
    \label{tab:reward_terms_for_training}
    \vspace{-4mm}
\end{table}

The observation of RL policy in this task includes the following elements:
\begin{itemize}
    \item The position and velocity of the six key points placed on the three corners of each leg soles with the base link as the origin (36 dims)
    \item Command in the previous step (20 dims)
    \item Body orientation in the Euler angle (3 dims)
    \item Body angular velocity (3 dims)
    \item Periodic phase signal (2 dims, only for walking)
\end{itemize}
To enable policies considering the time series of the state, we concatenate the above observation obtained in the previous $n$ step with the observation obtained in the latest step, then feed it to the policy.
We use $n=1$ for the balancing task and $n=3$ for the walking task.

The reward function is designed similarly to the prior study by \cite{Siekmann2021commongaits}, so refer to their paper for some details.
\footnotesize
\begin{eqnarray}
    R(s, p) & = & K_{bipedal} \cdot R_{bipedal}(s, p) + K_{cmd} \cdot R_{cmd}(s)  \nonumber\\ 
            & + & K_{smooth} \cdot R_{smt}(s) + 1 \nonumber\\
            \nonumber\\
    R_{cmd}(s) & = & (-1) \cdot (1 - exp(-K_{xd} \cdot | \dot{x}_{desired} - \dot{x}_{actual} |)  \nonumber\\
                & + & (-1) \cdot (1 - exp(-K_{yd} \cdot | \dot{y}_{desired} - \dot{y}_{actual} |) \nonumber\\
    \label{reward_function}   & + & (-1) \cdot (1 - exp(-4 \cdot |r_{desired} - r_{actual}|) \\
                \nonumber\\
    R_{smt}(s) & = &(-1) \cdot (1 - exp(-0.1 \cdot |a_t - a_{t-1}|)) \nonumber\\  
               & + & (-1) \cdot (1 - exp(-0.05 \cdot | \mathbf{I} \cdot 10 |) \nonumber\\
               & + & (-1) \cdot (1 - exp(-0.1 \cdot (| \dot{r} | + | acc_{pelvis} |)))\nonumber
\end{eqnarray}
\normalsize
Where, $R_{bipedal}$ is the penalty term for learning to step by the leg synchronized with the periodic signal by eq.~(2) in \cite{Siekmann2021commongaits}.
Each coefficient $K$ has a different value appropriate for each task as shown in Table~\ref{tab:reward_terms_for_training}.
For example, since stepping is unnecessary for the balancing task, $K_{bipedal}$ was set to zero.
$p$ is the periodic phase signal only used for the walking task.
$\dot{x}$ and $\dot{y}$ are the translation velocity of robot body, $r$ is body orientation, $\dot{r}$ is body angular velocity, $acc_{pelvis}$ is a linear acceleration of robot body.
$\mathbf{I}$ is the vector of joint motor currents, and $a_t$ is the action vector commanded by policy at time step $t$.

We use eq.~\eqref{reward_function} for $R_{eval}$ in the re-identification process with coefficients on Table~\ref{tab:reward_terms_for_training}.
Note that due to the sensing limitation on the actual robot setting, $\dot{x}_{actual}$ and $\dot{y}_{actual}$ are approximately estimated. Details are in each task section.

\subsubsection{Balancing task}
\label{sec:balancing task}
The objective of the balancing task is to maintain balance during standing upright on a board with a dynamically changing tilt.
Since the center of gravity movement should be minimized, $\dot{x}_{desired}$ and $\dot{y}_{desired}$ on the reward are always set to zero.
During training, a board on which the robot stands vibrates with a randomly generated wave in the roll-pitch axis.
The wave amplitude is up to $10^\circ$.
Training episodes are terminated when the upper body tilts largely or the body height lower the threshold.
For evaluation of the actual robot, we use a board that tilts $6^\circ$ for the pitch axis.
With the unknown board angle, the control policy quickly tries to get the upper body posture vertical and stationary.
To calculate $\dot{x}_{actual}$ and $\dot{y}_{actual}$ on $R_{eval}$, we use approximated value from body angular velocities.

\subsubsection{Walking task}
\label{sec:Walking task}
The bipedal walking task is designed as a task that requires more active motion than the balancing task.
The objective is to walk forward on a flat surface at a constant speed.
The target walking cycle was set at $1.0~sec$, and the ratio of the support leg phase to the swing leg phase was set at $0.7:0.3$.
We set the target speed as $\dot{x}_{desired}=0.3$ and $\dot{y}_{desired}=0$.
For evaluation of the actual robot, we make it attempt to walk on a flat board.
The position where the inner tip of the robot foot traveled furthest in the forward direction is recorded and divided by time to obtain the average speed, which is used as $\dot{x}_{actual}$ and $\dot{y}_{actual}$ in $R_{eval}$ at each time step.
\subsection{DRL training and optimization setup}
\label{sec:computation}
For the experiments, including DRL training and black-box optimizations in the system identification process, we use a machine that equips Intel Core i7-9700KF CPU @ $3.60GHz$ and NVIDIA GeForce RTX 2060 GPU.
The neural network of the policy of DRL is fully-connected and has two hidden layers with ReLU activation.
The node size of each hidden layer is set to 256 for the balancing task and 1024 for the walking task.
We train a Soft-Actor-Critic agent for $1,000,000$ steps for the balancing task and $3,000,000$ steps for the walking task, with a training time of approximately 5 and 15 hours, respectively.
Black-box optimizations in system identification take about 2 hours for the first system identification and about 5 hours for re-identification.

\begin{figure}[t]
    \centering
    \includegraphics[keepaspectratio, scale=0.4]{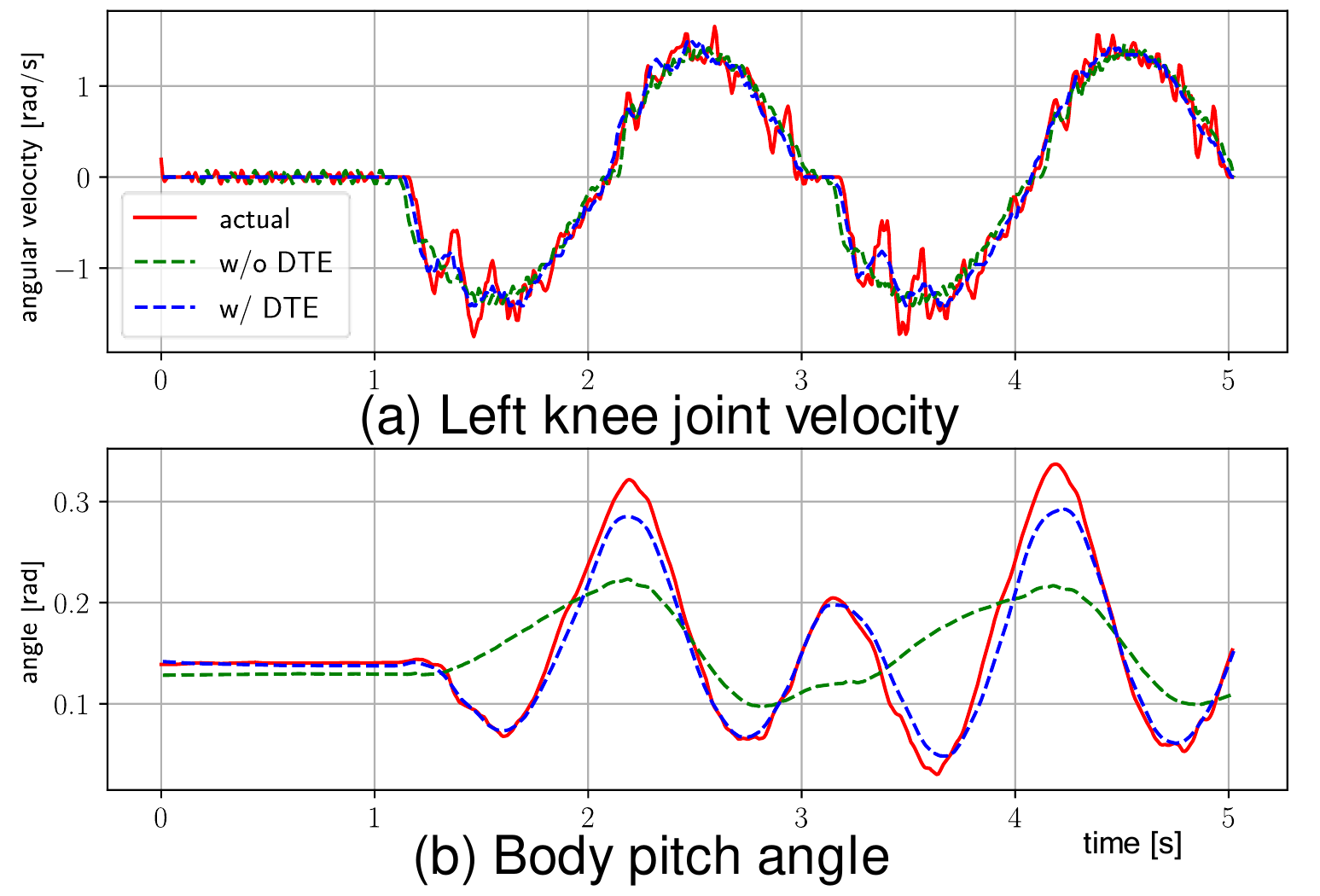}
    \caption{Result of system identification with squat motion}
    \label{fig:sim_calib_result}
    \vspace{-2mm}
\end{figure}
\section{Results}
\label{sec:result}
The following three points are evaluated to verify the effectiveness of the proposed method:
1) the effectiveness of the actuator model for sim-to-real (\Cref{sec:Evaluation of Actuator Modeling}), 2) the proposed system identification method for sim-to-real (\Cref{sec:Evaluation of model identification method}), and 3) how well does the trained policy perform on the actual robot (\Cref{sec:Policy Performance}).
\subsection{Evaluation of Actuator Modeling}
\label{sec:Evaluation of Actuator Modeling}
First, we show the effect of the DTE model in the first system identification with the squatting motion.
We performed the optimization of eq.~\eqref{eq:sim_calib_overview} with and without the DTE model, respectively.
The error evaluation score according to eq.~\eqref{eq:IdenLossFunc} was 0.037 when the DTE model was used (``w/ DTE'') and 0.070 when it was not used (``w/o DTE'').
Fig.~\ref{fig:sim_calib_result} shows the graphs of the sensor data of the actual and each simulated robot results.
``w/ DTE'' model reproduced sensor data better in comparison to ``w/o DTE''.
In particular, the DTE's ability to represent changes in a frictional loss in static and driving conditions is thought to have contributed to the reproducibility of the overall behavior.

\subsection{Evaluation of Sim-to-real Transfer Method}
\label{sec:Evaluation of model identification method}
In this section, we evaluate the proposed system identification method for sim-to-real transfer.
We performed the sim-to-real transfer process described in \Cref{sec:proposed_method} for the balancing and walking tasks.
The whole transfer process was conducted three times, with different seed values, and the results are shown in Table~\ref{tab:sim_to_real_results} and Table~\ref{tab:sim_to_real_walking}.
The left column lists the names of the set of simulation parameters $\phi$ to train the policy, and the right column shows the number of successful transfer attempts.
``Excitation only'' corresponds to the identified simulation parameters with the first system identification at phase 1, and ``Re-Identified'' corresponds to the re-identified parameters at phase 3.
For the balancing task, the success of the transfer was determined by comparing the rewards obtained in the simulation and on the actual robot.
The policy was run ten times in the evaluation setting on the simulation with parameter $\phi$ used for policy training and evaluated with $R_{eval}$ to obtain the expected reward value.
Then, the policy was run ten times on the actual robot, and the transfer was judged successful if the average reward value achieved more than 80\% of the expected value.
For the walking task, we executed each policy ten times on the actual robot and judged the transfer successful when the robot could go forward without falling for ten seconds in eight episodes.

\begin{table}[t]
    \centering
    \caption{Success rates of sim-to-real transfer in balancing task}
    \begingroup
    \scalefont{0.85}
    \begin{tabular}{c||c}
        \hline
        \begin{tabular}{c}
          Parameter used for\\policy training
        \end{tabular} &
        \begin{tabular}{c}
          Balancing on\\tilting board
        \end{tabular} \\
        \hline
        \hline
        Excitation only             & 0/3  \\
        Re-Identified (ours)        & \bf{3/3} \\
        Excitation only (w/o DTE)   & 0/3 \\
        Re-Identified (w/o DTE)     & 0/3 \\
        Domain randomization        & 0/3 \\
    \end{tabular}
    \endgroup
    \label{tab:sim_to_real_results}
    \vspace{-2mm}
\end{table}

\begin{table}[t]
\centering
\caption{Success rates of sim-to-real transfer in walking task}
\begin{threeparttable}
    \begingroup
    \scalefont{0.85}
        \begin{tabular}{c||c}
            \hline
            \begin{tabular}{c}
              Parameter used for\\policy training
            \end{tabular} &
            Walking on a flat floor \\
            \hline\hline
            Excitation only      & 0/3 \\
            Re-Identified (ours) & \bf{3/3} \\
            Domain randomization & 0/3 \\
        \end{tabular}
    \endgroup
    \end{threeparttable}
    \label{tab:sim_to_real_walking}
    \vspace{-4mm}
\end{table}

As shown in Table~\ref{tab:sim_to_real_results} and Table~\ref{tab:sim_to_real_walking}, policies trained with ``Excitation only'' failed to work on the actual robot.
Then re-identification was processed, and policies trained with ``Re-Identified'' successfully transferred in all attempts.
Despite the small amount of actual robot operation time for data collection (only 5 seconds of squat motion and ten episodes on the actual robot), sim-to-real transfer was achieved without domain randomization.
``Excitation only (w/o DTE)'' and ``Re-Identified (w/o DTE)'' in Table~\ref{tab:sim_to_real_results} shows the results of when we performed the sim-to-real transfer process using simulation with the DTE model excluded.
As shown in the table, it failed to sim-to-real without the DTE model.
The ``Re-Identified (w/o DTE)'' made little difference to the behavior of the failure, and the policy fell over because it could not control its movements.
The re-identification process found parameters that reduced the policy performance gap while limiting the increase in the $L_{exc}$ but was still considered far from the actual behavior due to the limited reproducibility seen in Fig.~\ref{fig:sim_calib_result}.
From this result, we can say that DTE is one of the critical components of the actuator model in our experimental settings.
As the need for a DTE model was demonstrated, experiments with a walking task were omitted.

\begin{figure}[t]
    \begin{minipage}[b]{\linewidth}
    \centering
    \includegraphics[width=0.85\columnwidth]
    {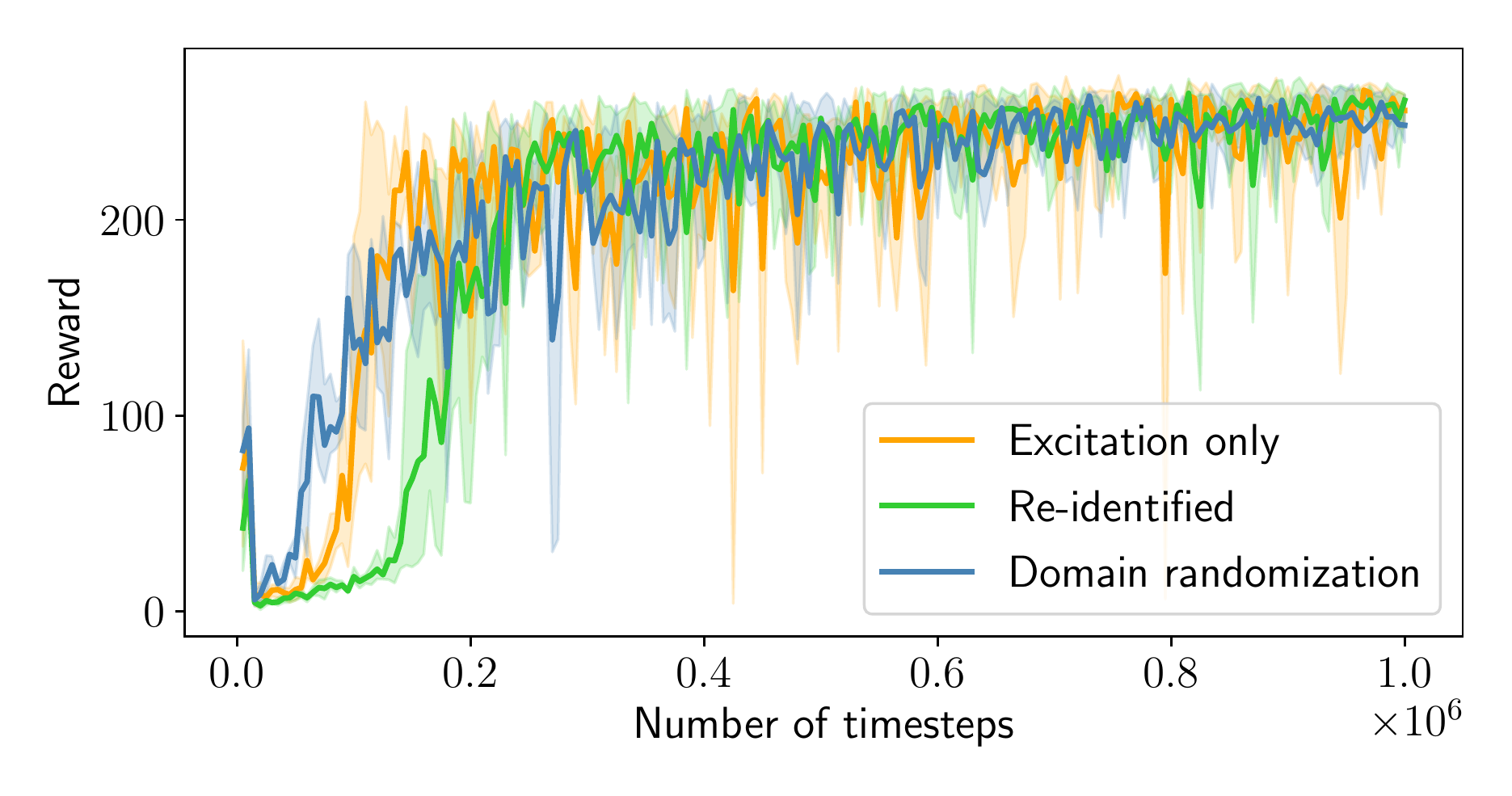}
    \end{minipage}
    \caption{Learning curves of RL training in balancing task}
    \label{fig:learning_curve}
    \vspace{-4mm}
\end{figure}

In addition, we trained policies with naive domain randomization as a baseline.
During training, each simulation parameter is uniformly sampled within a range shown in Table~\ref{tab:model_paramters}.
We conducted the training with three seed values, but the transferred policy could not perform well on the actual robot in any attempt.
Fig.~\ref{fig:learning_curve} shows learning curves in the balancing task, showing that the average performance on the randomly sampled parameter sets converged.
This observation suggests that policies that work well in most parameter sets were well-trained, but they didn't work in some local parameter sets, including those that are close to the real.
Such a problem requires a process that can learn appropriate actions for parameter sets close to the actual robot, as in the proposed method.

\begin{figure}[t]
    \centering
    \includegraphics[width=1.0\columnwidth]{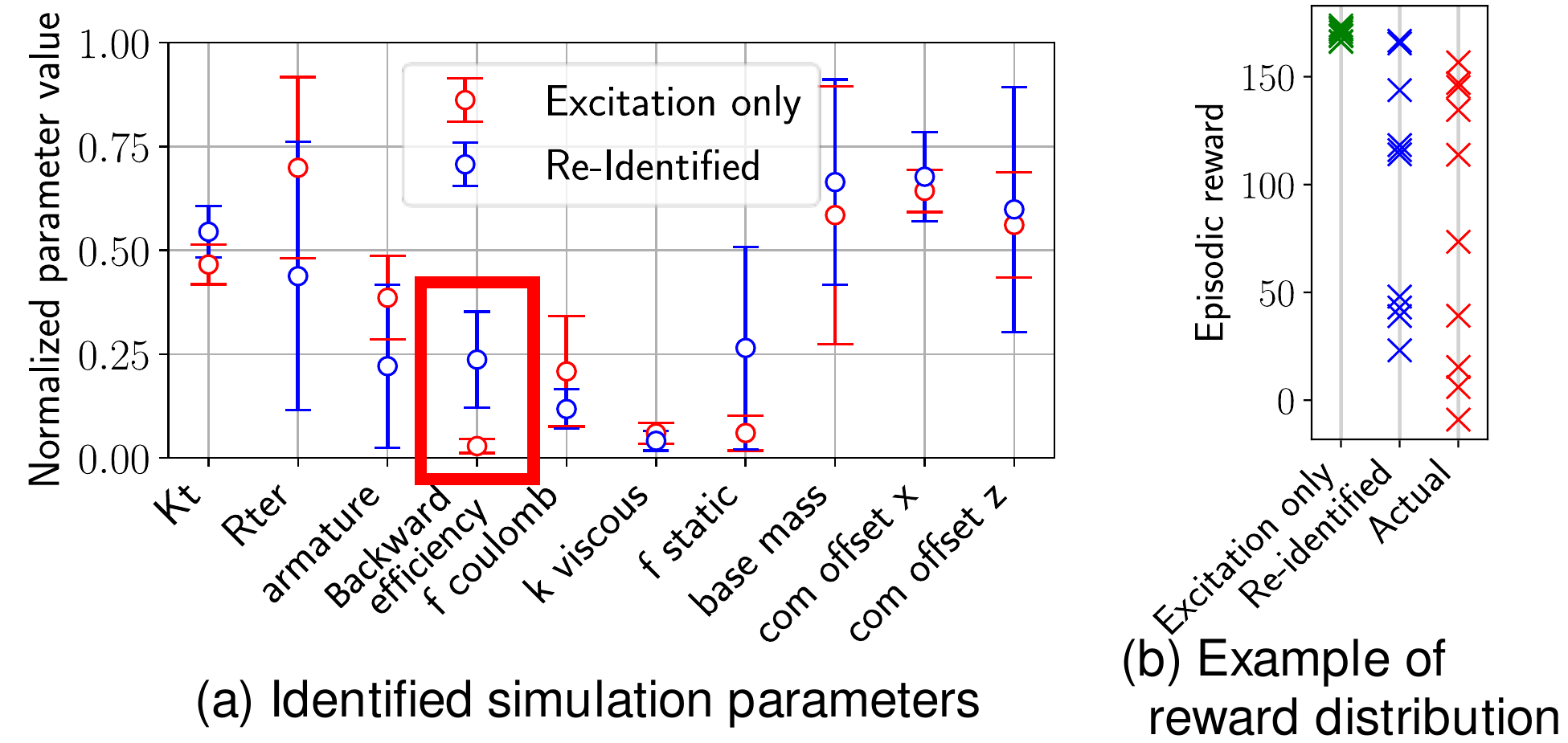}
    \caption{Materials for analysis of re-identification process.}
    \label{fig:identified_simulation_parameters}
    \vspace{-4mm}
\end{figure}

We analyzed the results of the reward value and identified simulation parameters.
Fig.~\ref{fig:identified_simulation_parameters}~(b) shows an example of reward distribution from the walking task experiment.
In the graph, ``Excitation only'' shows reward distribution on the simulation used for training, ``Re-Identified'' shows rewards acquired by the same policy on the simulation after re-identification, and ``Actual'' shows rewards on the actual.
We can see re-identification could find the simulation parameters to close the reward distribution actual.
If the reward distribution is close between the actual robot and the simulation, similar behavior can be produced.
Therefore, we can say that the re-identification approach finds the simulation parameters closer to the actual.
Fig.~\ref{fig:identified_simulation_parameters}~(a) shows the mean and standard deviation of identified simulation parameter values of both the balancing and walking tasks.
Values are normalized by the range shown in \Cref{tab:model_paramters}.
The variances are not slight, but this is likely due to several combinations of parameter values showing similar physical behavior.
Among them, backward efficiency tends to increase with re-identification.
In both tasks, leg joint oscillations and falls were observed in actual cases of failed transfers, apparently due to the joints not stopping due to reaction forces from the ground.
Parameter changes are likely to be the result of re-identification reflecting these data.


\begin{figure}[t]
    \begin{minipage}[b]{\linewidth}
    \centering
    \includegraphics[keepaspectratio, scale=0.2]{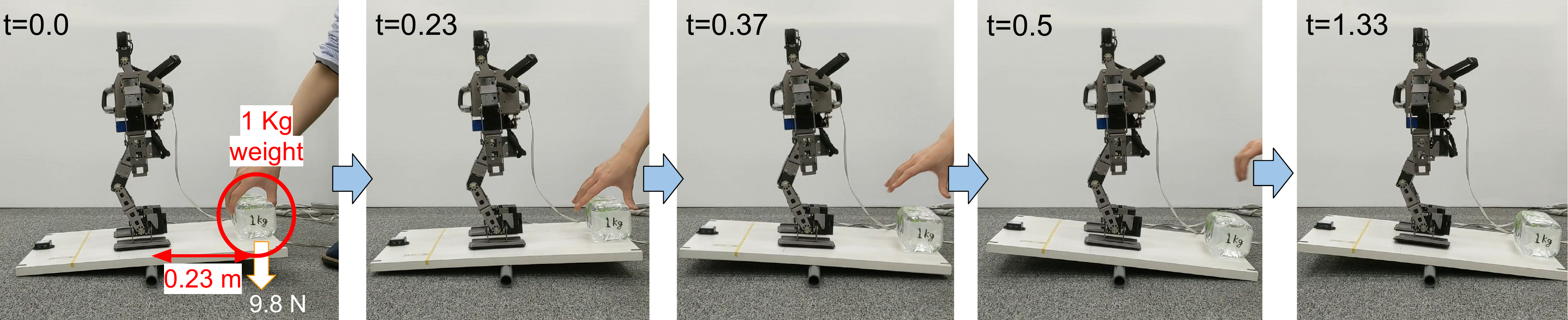}
    \end{minipage}
    
    \caption{Snapshots of balancing behavior. Applying disturbance by putting the weight.}
    \label{fig:balancing_on_uneven_surface_board}
    \vspace{-4mm}
\end{figure}
\subsection{Evaluation of Policy Performance}
\label{sec:Policy Performance}
To verify the performance of the policy acquired with the proposed method, we demonstrate and evaluate the tasks on an actual robot.
Each motion of the policy can be viewed in the supplementary video.

\subsubsection{Balancing Task}
Fig.~\ref{fig:balancing_on_uneven_surface_board} shows the snapshot of the balancing task experiment.
The robot is placed on a board that freely tilts from $-6^\circ$ to $6^\circ$ on the pitch axis.
Disturbances are provided by gently placing a $1\mathrm{\,Kg}$ weight approximately $0.23\mathrm{\,m}$ from the center of rotation.
It was confirmed that the policy could maintain balance under large disturbances, and the robot could maintain an upright posture.

\begin{figure}[t]
    \begin{minipage}[b]{\linewidth}
    \centering
    \vspace{1mm}
    \includegraphics[width=0.725\columnwidth]
    {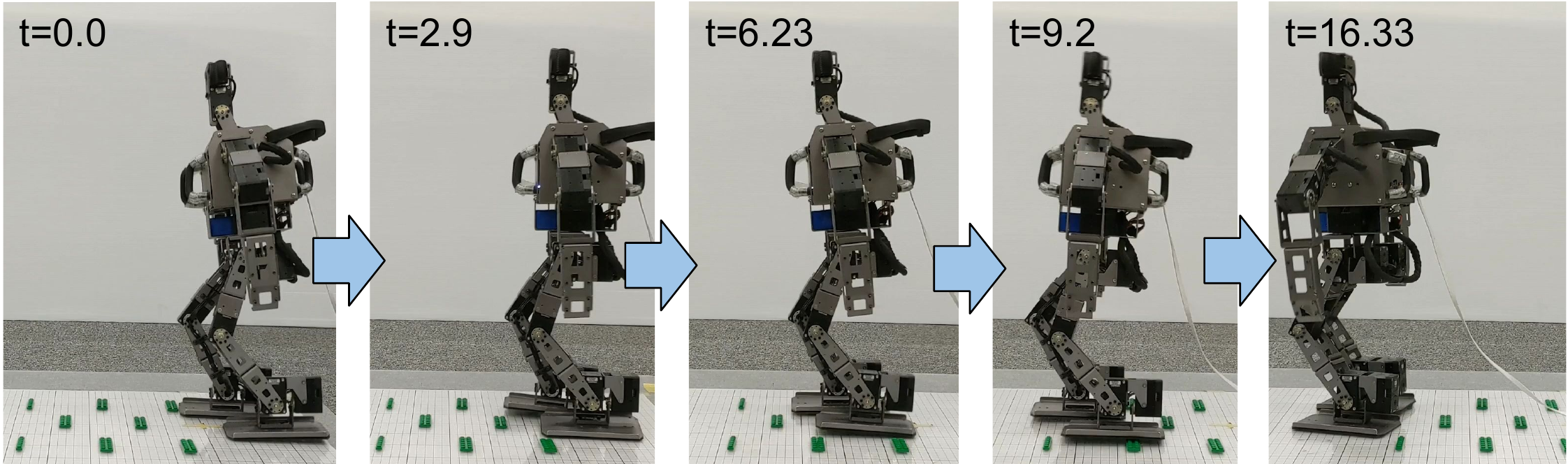}
    \subcaption{Snapshots of walking behavior}
    \end{minipage}
    \begin{minipage}[b]{\linewidth}
    \centering
    \vspace{1mm}
    \includegraphics[width=0.725\columnwidth]
    {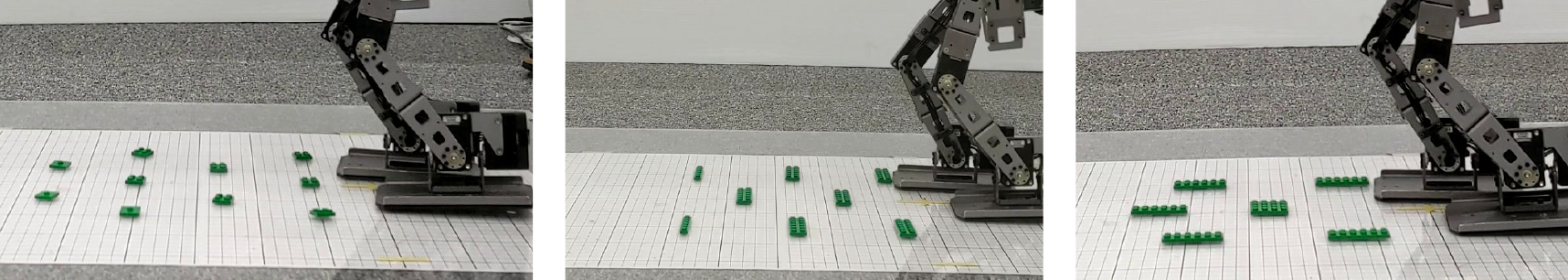}
    \subcaption{Tested uneven surface patterns}
    \end{minipage}
    \caption{Snapshots of walking experiments on uneven surfaces}
    \label{fig:walking}
    \vspace{-4mm}
\end{figure}

\begin{figure}[t]
    \begin{minipage}[b]{\linewidth}
    \centering
    \includegraphics[width=0.825\columnwidth]{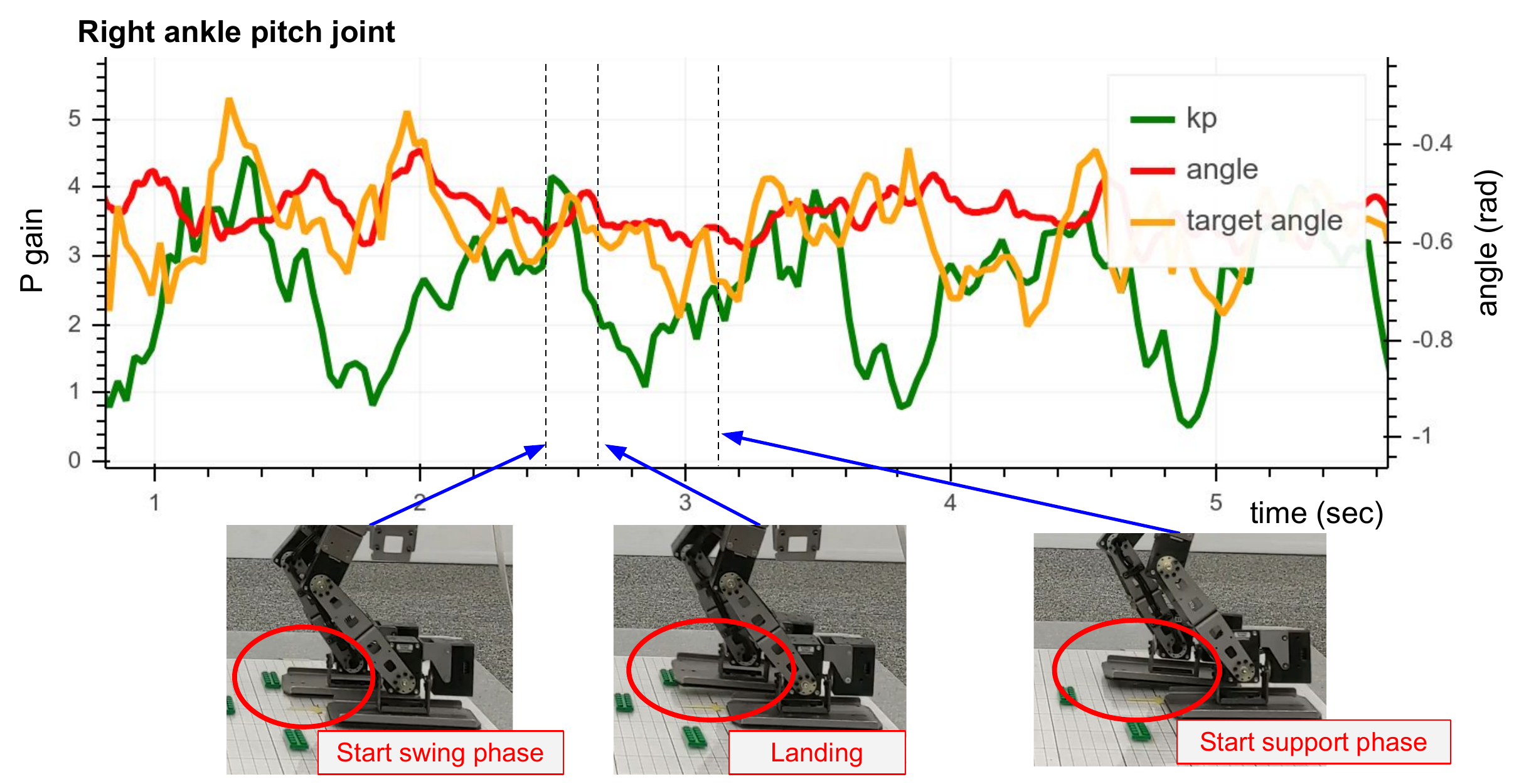}
    \end{minipage}
    
    \caption{Policy's output command while walking}
    \label{fig:policy_command}
    \vspace{-5mm}
\end{figure}

\subsubsection{Walking Task}
\label{sec:Walking Experiment}
To realize the robust policy utilizing the backdrivability this research aims for, we trained a walking policy on the simulation with random uneven terrains.
For training, the re-identified simulation parameters with the walking task were used.
The trained policy passed three uneven surfaces for testing; the surface patterns and snapshots of walking are shown in Fig.~\ref{fig:walking}.
We used some toy brick parts to introduce various unevenness.
Each brick has a $5\mathrm{\,mm}$ height on top.
Fig.~\ref{fig:policy_command} shows the right ankle pitch joint command by the policy while walking.
The P gain is reduced as the leg transitions into the support phase, indicating a compliant landing.

In addition, we tested an existing walking module of ROBOTIS-OP3 with the default settings as a baseline using stiff position-controlled joints. The walking module failed on these uneven surfaces.
The fact that the policy could traverse uneven surfaces, which is difficult with a general walking controller, without force or contact sensors indicates that the policy can take advantage of backdrivability.


\section{Discussion}
\label{sec:Discussion}

The trained policy demonstrated robustness to uneven terrain, but the motion smoothness and walking speed are future issues.
One of the main reasons for the low policy performance is the PD control of the joints.
In this paper, we did not use DYNAMIXEL's internal position control.
As mentioned in \Cref{sec:robot setup}, this is to avoid performing a system identification that includes the unknown behavior of DYNAMIXEL's position control that it may have.
Instead, we used a system that calculates PD control by PC inside the robot.
However, in this system, due to the communication limitation, the control frequency of the PD control is 125 Hz, which is much lower than general, and there is command latency.
Due to these factors, we could not ensure the position control's stability if the D-gain was increased beyond a certain level, so the D-gain had to be set excessively low.
This is also why the D-gain change was omitted from the policy action.
The low D-gain made the position control behave like a spring and was difficult for the policy to handle, leading to poor performance.
A possible way to improve policy performance would be to utilize DYNAMIXEL's internal position control.
DYNAMIXEL's internal position control is better, at least in terms of control frequency and latency, and it is expected that D-gain could also be added to the policy's actions. 
To achieve sim-to-real after replacement, a system identification including the DYNAMIXEL's position control would have to be carried out while utilizing the results of this paper.

\section{Conclusion}
\label{sec:conclusion}
In this paper, we proposed a sim-to-real transfer method featuring the actuator model with the DTE and a re-identification method that utilizes the acquired rewards of failed transfer.
The actuator model improved the ability to reproduce the robot motion, and the re-identification method enabled the identification of more realistic parameters that reproduce the failed behavior in the actual robot, resulting in a successful transfer.
The method was verified on the torque sensor-less gear-driven robot ROBOTIS-OP3 and achieved balancing under disturbance and walking on uneven surfaces.

\section*{Acknowledgement}
The authors thank Yasuhiro Fujita and Avinash Ummadisingu for their support.

\bibliographystyle{IEEEtran} 
\bibliography{bibliography}

\begin{thebibliography}{10}
\providecommand{\url}[1]{#1}
\csname url@rmstyle\endcsname
\providecommand{\newblock}{\relax}
\providecommand{\bibinfo}[2]{#2}
\providecommand\BIBentrySTDinterwordspacing{\spaceskip=0pt\relax}
\providecommand\BIBentryALTinterwordstretchfactor{4}
\providecommand\BIBentryALTinterwordspacing{\spaceskip=\fontdimen2\font plus
\BIBentryALTinterwordstretchfactor\fontdimen3\font minus
  \fontdimen4\font\relax}
\providecommand\BIBforeignlanguage[2]{{%
\expandafter\ifx\csname l@#1\endcsname\relax
\typeout{** WARNING: IEEEtran.bst: No hyphenation pattern has been}%
\typeout{** loaded for the language `#1'. Using the pattern for}%
\typeout{** the default language instead.}%
\else
\language=\csname l@#1\endcsname
\fi
#2}}

\bibitem{Lee2020quadwild}
\BIBentryALTinterwordspacing
J.~Lee, \emph{et~al.}, ``Learning quadrupedal locomotion over challenging
  terrain,'' \emph{Science Robotics}, vol.~5, no.~47, Oct 2020. [Online].
  Available: \url{http://dx.doi.org/10.1126/scirobotics.abc5986}
\BIBentrySTDinterwordspacing

\bibitem{Siekmann2021cassie-blind}
J.~Siekmann, \emph{et~al.}, ``Blind bipedal stair traversal via sim-to-real
  reinforcement learning,'' https://arxiv.org/abs/2105.08328, 2021.

\bibitem{Peng2020imitation}
X.~B. Peng, \emph{et~al.}, ``Learning agile robotic locomotion skills by
  imitating animals,'' in \emph{Robotics: Science and Systems}, 07 2020.

\bibitem{tan2018simtoreal}
\BIBentryALTinterwordspacing
J.~Tan, \emph{et~al.}, ``Sim-to-real: Learning agile locomotion for quadruped
  robots,'' 2018. [Online]. Available: \url{https://arxiv.org/abs/1804.10332}
\BIBentrySTDinterwordspacing

\bibitem{du2021autotuned}
Y.~Du, \emph{et~al.}, ``Auto-tuned sim-to-real transfer,'' in \emph{2021 IEEE
  International Conference on Robotics and Automation (ICRA)}, 2021, pp.
  1290--1296.

\bibitem{Yu2019PUP}
W.~Yu, \emph{et~al.}, ``Sim-to-real transfer for biped locomotion,'' in
  \emph{2019 IEEE/RSJ International Conference on Intelligent Robots and
  Systems (IROS)}, 2019, pp. 3503--3510.

\bibitem{tobin2017domain}
J.~Tobin, \emph{et~al.}, ``Domain randomization for transferring deep neural
  networks from simulation to the real world,''
  https://arxiv.org/abs/1703.06907, 2017.

\bibitem{chebotar2019closing}
Y.~Chebotar, \emph{et~al.}, ``Closing the sim-to-real loop: Adapting simulation
  randomization with real world experience,'' in \emph{2019 International
  Conference on Robotics and Automation (ICRA)}, 2019, pp. 8973--8979.

\bibitem{Xie2020cassie-simtoreal}
Z.~Xie, \emph{et~al.}, ``Learning locomotion skills for cassie: Iterative
  design and sim-to-real,'' in \emph{Proceedings of the Conference on Robot
  Learning}, ser. Proceedings of Machine Learning Research, vol. 100, 2020, pp.
  317--329.

\bibitem{Hwangbo2019actuatornet}
\BIBentryALTinterwordspacing
J.~Hwangbo, \emph{et~al.}, ``Learning agile and dynamic motor skills for legged
  robots,'' \emph{Science Robotics}, vol.~4, no.~26, Jan 2019. [Online].
  Available: \url{http://dx.doi.org/10.1126/scirobotics.aau5872}
\BIBentrySTDinterwordspacing

\bibitem{albert2015directional}
A.~Wang and S.~Kim, ``Directional efficiency in geared transmissions:
  Characterization of backdrivability towards improved proprioceptive
  control,'' in \emph{2015 IEEE International Conference on Robotics and
  Automation (ICRA)}, 2015, pp. 1055--1062.

\bibitem{Matsuki2019backdrive}
H.~Matsuki, \emph{et~al.}, ``Bilateral drive gear—a highly backdrivable
  reduction gearbox for robotic actuators,'' \emph{IEEE/ASME Transactions on
  Mechatronics}, vol.~24, no.~6, pp. 2661--2673, 2019.

\bibitem{Hyon2006passivity}
S.~Hyon and G.~Cheng, ``Passivity-based full-body force control for humanoids
  and application to dynamic balancing and locomotion,'' in \emph{2006 IEEE/RSJ
  International Conference on Intelligent Robots and Systems}, 2006, pp.
  4915--4922.

\bibitem{Mesesan2019passivity}
G.~Mesesan, \emph{et~al.}, ``Dynamic walking on compliant and uneven terrain
  using dcm and passivity-based whole-body control,'' in \emph{2019 IEEE-RAS
  19th International Conference on Humanoid Robots (Humanoids)}, 2019, pp.
  25--32.

\bibitem{Suzuki2017sensorless-torque}
H.~Suzuki, \emph{et~al.}, ``Torque based stabilization control for torque
  sensorless humanoid robots,'' in \emph{2017 IEEE-RAS 17th International
  Conference on Humanoid Robotics (Humanoids)}, 2017, pp. 425--431.

\bibitem{Siekmann2021commongaits}
J.~Siekmann, \emph{et~al.}, ``Sim-to-real learning of all common bipedal gaits
  via periodic reward composition,'' in \emph{2021 IEEE International
  Conference on Robotics and Automation (ICRA)}, 2021, pp. 7309--7315.

\bibitem{Peng2017actionspace}
X.~B. Peng and M.~van~de Panne, ``Learning locomotion skills using deeprl: Does
  the choice of action space matter?'' in \emph{Proceedings of the ACM SIGGRAPH
  / Eurographics Symposium on Computer Animation}, ser. SCA '17, 2017, pp.
  12:1--12:13.

\bibitem{taylor2021sonybipedal}
M.~Taylor, \emph{et~al.}, ``Learning bipedal robot locomotion from human
  movement,'' in \emph{2021 IEEE International Conference on Robotics and
  Automation (ICRA)}, 2021, pp. 2797--2803.

\bibitem{haarnoja2018sac}
T.~Haarnoja, \emph{et~al.}, ``Soft actor-critic: Off-policy maximum entropy
  deep reinforcement learning with a stochastic actor,''
  https://arxiv.org/abs/1801.01290, 2018.

\bibitem{Todorov2012mujoco}
E.~Todorov, \emph{et~al.}, ``Mujoco: A physics engine for model-based
  control,'' in \emph{2012 IEEE/RSJ International Conference on Intelligent
  Robots and Systems}, 2012, pp. 5026--5033.

\bibitem{tan2016simcalib}
J.~Tan, \emph{et~al.}, ``Simulation-based design of dynamic controllers for
  humanoid balancing,'' in \emph{2016 IEEE/RSJ International Conference on
  Intelligent Robots and Systems (IROS)}, 2016, pp. 2729--2736.

\bibitem{ROBOTIS-OP3}
{ROBOTIS CO.,LTD}, ``{ROBOTIS-OP3 INTRODUCTION},''
  \url{https://emanual.robotis.com/docs/en/platform/op3/introduction/}(accessed
  on 28 Feb 2023).

\bibitem{Akiba2019optuna}
T.~Akiba, \emph{et~al.}, ``Optuna: A next-generation hyperparameter
  optimization framework,'' in \emph{Proceedings of the 25rd {ACM} {SIGKDD}
  International Conference on Knowledge Discovery and Data Mining}, 2019.

\end{thebibliography}
\end{document}